\pdfoutput=1
\documentclass[runningheads]{llncs}

\usepackage{eccv}
\usepackage[width=122mm,left=13mm,paperwidth=146mm,height=193mm,top=12mm,paperheight=217mm]{geometry}

\usepackage{eccvabbrv}
\usepackage{graphicx}
\usepackage{xcolor}
\usepackage{booktabs}
\usepackage{amsmath}
\usepackage{amssymb}
\usepackage{multirow}
\usepackage{caption}
\usepackage[accsupp]{axessibility}

\usepackage{hyperref}
\usepackage{orcidlink}

\graphicspath{{figures/}}

\begin{document}

\title{Multi-Year Geospatial Reasoning using Interannually-Consistent
   Historical Predictions as a Free Input Modality}

\titlerunning{Historical Predictions as a Free Input Modality}

\author{Syed Roshaan Ali Shah\orcidlink{0000-0001-9517-5142} \and Kasper Bonte\orcidlink{0000-0002-0221-9458} \and David Bekaert \and
Kristof Van Tricht\orcidlink{0000-0001-5727-9098} \and Dieter Wens}
\authorrunning{Shah \, S.R.A. et al.}
\institute{VITO Remote Sensing, Mol, Belgium\\
}

\maketitle

\vspace{-12pt}
\begin{abstract}

Machine learning, and deep networks in particular, are increasingly used to derive higher-level Earth observation (EO) products such as annual land-cover and crop-type maps. Many are generated operationally: each year a new acquisition is processed, typically with the same model, extending a multi-year archive. In the process these systems accumulate two kinds of useful signal that are almost never fed back into the model: the system's own archive of past predictions, and ancillary layers produced by other partners in a processing consortium. Both are normally used outside the network, as rule-based post-processing or a fixed input mask. Using the Copernicus Land Monitoring Service High Resolution Layer (HRL) Croplands crop-type product \cite{hrlpum2025} as a testbed, we show that bringing both signals inside the model turns a single-year, single-task pixel classifier into one that reasons across years. We introduce a Crop Type (CTY) embedding encoder that represents each past prediction as a confidence-scaled, time-ordered categorical token and attends over the year axis, and we study how the externally provided Base Vegetation Layer (BVL) mask should be represented in the model's inputs and outputs. To compare designs fairly when they relabel non-crop pixels, we evaluate on the 18 crop classes only and report precision and recall separately. On a pan-European dataset of about 5.4M labelled pixels, adding the prediction history raises crop-only F1 by 1.6 percentage points (pp) and, more importantly, corrects a recall-skewed error profile, with the largest gains on perennial and tree crops (olives +4.6, fruits +3.7, nuts +3.2 pp). Representing the BVL mask consistently in both the history and the target year adds about 2.5 pp on the crop classes. The approach is a low-cost recipe for any recurring geospatial or foundation model that emits class maps.

\end{abstract}

\section{Introduction}
\label{sec:intro}

High-resolution maps of crop types support agricultural monitoring, food
security assessment, and policy frameworks such as the European Common
Agricultural Policy. Such maps are increasingly produced operationally at
continental scale and 10\,m resolution using deep networks over Sentinel
satellite image time series
\cite{dandrimont2021parcel,vantricht2023worldcereal,brown2022dynamicworld},
with many products updated annually. One property of these systems is
that they are recurring, re-running essentially the same or slightly updated model
each production cycle. A common limitation follows: although the service
accumulates a dense, per-pixel record of its own past predictions, that
record is never fed back into the model.

Despite this, the standard approach classifies each year on its own. The
map for 2024 sees only the 2024 time series and has no access to the fact
that the same pixel was olive from 2017 to 2023. Where multi-year
consistency is enforced at all, it is added after inference as rule-based
post-processing \cite{quinton2021crop}. A second, related limitation
appears in consortium-based production. An ancillary Base Vegetation
Layer (BVL) mask, produced by one partner, is applied to the product
outside the model to remove non-crop pixels during post-processing. The
model never incorporates this signal, so when it later reads its own history, a pixel the BVL removed in some past years has no defined meaning
to it: it is neither a crop class nor plain missing data. 

Our position is that a recurring EO model can use what it already produces. It should read its own prediction history as an input, and it should represent the ancillary layers it consumes consistently inside the model, instead of treating either only as an out-of-model step. This turns a narrow production question into a
question about multi-year geospatial reasoning: can a model learn the
temporal grammar of land use, such as rotations, perennial persistence,
and gradual change, directly from its own outputs, and can it fold
external context into a single self-contained predictor? 

We study this on the Copernicus HRL Croplands crop-type product
\cite{hrlpum2025}, which has mapped crops annually since 2017. Our
approach consists of the following:

\begin{enumerate}
\item We treat prediction history as an input. Two aligned
sequences, the product's own past class predictions and their
winning-class probability, become an extra input modality,
fused with the existing optical, SAR, and meteorological encoders at no
labelling cost (\cref{sec:method}).
\item We introduce the CTY Embedding Encoder, which maps each past class
code to a learnable token, scales it by the confidence of that
prediction, adds a time ordering, and attends over years with a
padding-aware Transformer (\cref{sec:exp-strategy}).
\item We study how the externally provided vegetation mask should be represented inside the model: how masked observations are treated in the
target year, and whether they are remapped in the history. A consistent
representation improves the crop classes, without altering the mask
itself (\cref{sec:exp-bvl}).
\item We add cheap crop-rotation scalars and a soft self-supervised
consistency loss, and check robustness (\cref{sec:exp-prod}).
\end{enumerate}

The gains concentrate where multi-year reasoning should help most, on
perennial and tree crops with stable histories. We argue the recipe applies to any annual or seasonal EO product that emits recurring class maps, potentially including geospatial foundation models.

\section{Related Work}
\label{sec:related}
\vspace{-3pt}
\paragraph{Deep learning on satellite image time series.}
Crop mapping from Sentinel time series is now dominated by deep
sequence models. Temporal convolutional networks \cite{pelletier2019tempcnn}
and recurrent models gave way to attention-based methods.
Self-attention was applied to raw optical sequences for crop
classification by Ru\ss{}wurm and K\"orner \cite{russwurm2020selfattention},
formalised for multivariate series by Zerveas \etal
\cite{zerveas2021transformer} after the original Transformer
\cite{vaswani2017attention}, and made lightweight for satellite series
by the temporal attention encoders of Sainte Fare Garnot and Landrieu
\cite{garnot2020ltae,garnot2021utae}. Benchmarks such as BreizhCrops
\cite{russwurm2020breizhcrops} standardised evaluation. Our backbone
sits in this family; the contribution is orthogonal to the backbone and
concerns what the model is allowed to read.

\vspace{-3pt}
\paragraph{Self-supervision and foundation models for EO.}
Self-supervised pre-training reduces the need for labels, from SITS-BERT
\cite{yuan2021sitsbert} on time series to Presto \cite{tseng2024presto},
and to larger geospatial foundation models such as SatMAE
\cite{cong2022satmae}, Prithvi \cite{jakubik2023prithvi}, TerraFM \cite{danish2025terrafm} and AnySat 
\cite{astruc2024anysat}, with crop-specific deployment now being studied
operationally \cite{butsko2025worldcereal}. These learn transferable representations of imagery and are most valuable in label-scarce settings
or for transfer to new tasks. We instead add a representation of the
product's own prediction history as an extra contextual input channel~\cite{alishah2024contextual}, which is complementary and could be attached to a foundation backbone in the same
way.

\vspace{-3pt}
\paragraph{Multi-year and rotation-aware mapping.}
Most deployed systems treat each year independently. Several lines of
work add multi-year context from \emph{reference} labels: Quinton and
Landrieu \cite{quinton2021crop} model crop rotation explicitly and report
large gains for rotational and perennial crops; Bailly \etal
\cite{bailly2018rotation} structure prediction with a conditional random
field over past {LPIS} vintages; and Giordano \etal
\cite{giordano2020rotation} penalise predicting the same crop as the
previous year. Work on cross-year transfer \cite{wang2023domain }
documents the domain shift between years. A separate line enforces
temporal consistency \emph{after} classification, most notably the
hidden-Markov post-processing of Abercrombie and Friedl
\cite{abercrombie2016hmm} used to stabilise the {MODIS} land-cover
product. Our setting differs on both counts: the multi-year signal is the
model's \emph{own past predictions} rather than reference labels, and it
is consumed \emph{as an input} rather than corrected afterwards. 
The closest conceptual
analogue is knowledge distillation \cite{hinton2015distilling}: the model distils, then improves on, the output of a legacy rule-based stage.
\vspace{-3pt}
\paragraph{Class imbalance.}
Crop datasets are class-imbalanced: a few common crops dominate while many crops are rare (a long-tailed distribution). Focal loss \cite{lin2017focal} and
class-balanced weighting by effective number of samples
\cite{cui2019classbalanced} are standard.
We handle the long tail with balanced batch sampling, and add a failure
case showing that adding inverse-frequency \emph{loss} weighting on top
of that sampling is harmful at the imbalance ratios typical of
continental crop maps.

\paragraph{Reference data.}
Pan-European training and evaluation usually rely on harmonised in-situ archives
such as LUCAS-derived parcels \cite{dandrimont2021parcel}, the EuroCrops taxonomy \cite{schneider2023eurocrops}, and the WorldCereal reference system \cite{vantricht2023worldcereal}, which we follow.

\section{System and Two Sources of Free Signal}
\label{sec:background}

We build on an operational Copernicus Land Monitoring Service (CLMS)
product in the High Resolution Layer (HRL) Croplands suite, which has
produced annual 10\,m crop-type (CTY) maps since
2017 \cite{hrlatbd2025,hrlpum2025}. The wider product and its processing
chain matter for our argument, so we describe both before naming the two
signals it under-uses.

\subsection{The HRL Croplands Product Suite}
The suite is more than one map. The crop-type map assigns each cropland
pixel one of about 17 classes in a hierarchical legend \cite{schneider2023eurocrops}, used here at level~2 (the 18 crop
classes plus a no-crop class), together with a per-pixel confidence layer (CTYCL). Alongside it, the service publishes cropping-pattern layers:
main season emergence, duration and harvest, bare-soil periods, secondary season emergence and duration along with fallow indicators, and multi-year summaries of cropping
intensity and crop diversity computed over a rolling three-year window.
Multi-year reasoning therefore already exists in the product, but it
lives in separate, hand-built derived layers and rolling-window rules,
all computed downstream of a model that only ever sees one year. We ask
whether that reasoning belongs inside the model instead.

The production model is a pixel-based multi-source Transformer. Per-pixel
Sentinel-2 optical bands (9),
Sentinel-1 terrain-corrected $\sigma^0$ SAR backscatter (VV/VH),
derived
spectral indices, and AgERA5 meteorology (each a 24-step, 10-day
composited series) are encoded by a shared Transformer, while static
terrain scalars pass through a dense encoder. The two embeddings are
concatenated and fed to an MLP head (\cref{fig:arch}). The class is the
argmax, except that pixels whose top probability is below 35\% are marked
``unclassified''.

\subsection{The Legacy Postprocessing Chain}
\label{sec:background-postproc}
\begin{figure}[t]
  \centering
  \includegraphics[width=0.8\linewidth]{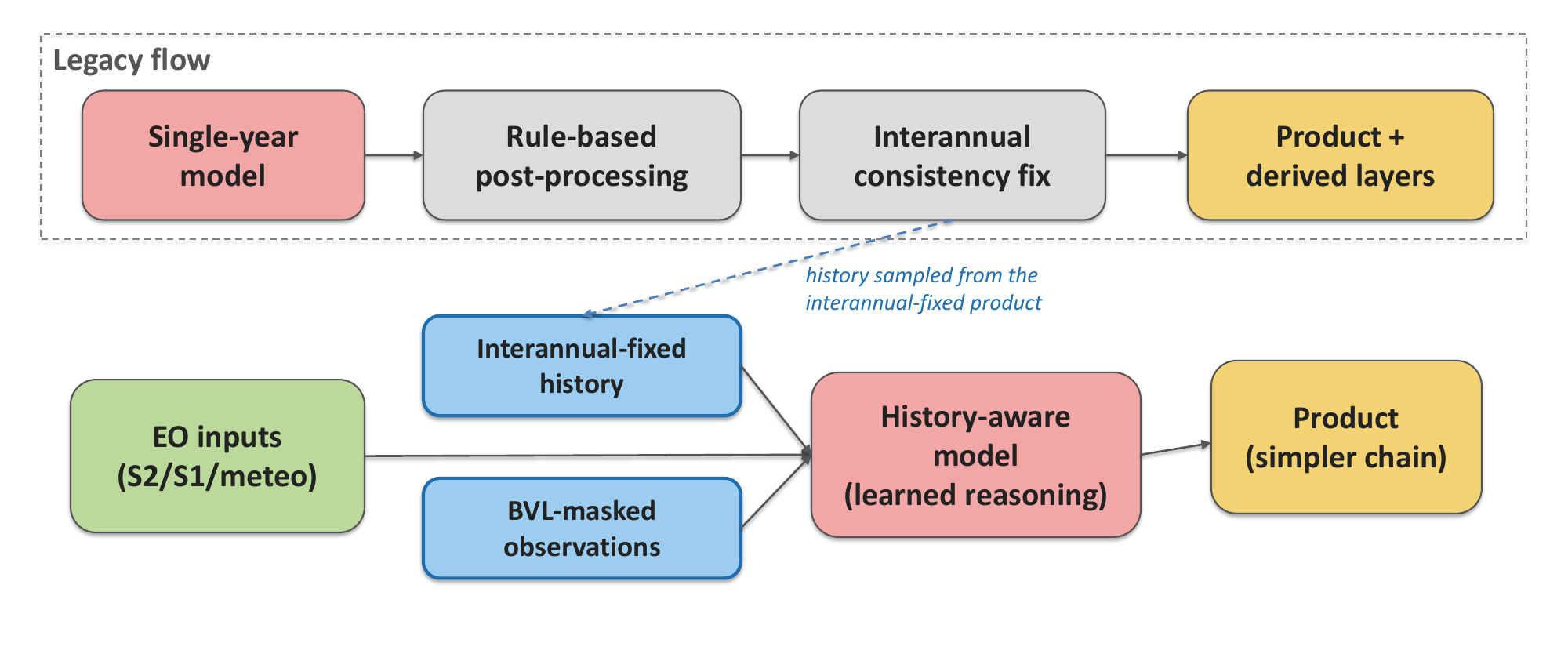}
  \caption{The operational chain. \emph{Top:} the legacy flow, a
  single-year model followed by rule-based post-processing, with
  multi-year reasoning living in separate derived layers. \emph{Bottom:}
  our version folds the interannual-fixed history and the BVL signal back
  into the model, so temporal reasoning is learned rather than applied
  afterwards.}
  \label{fig:pipeline}
  \vspace{-10pt}
\end{figure}

The raw yearly map is not shipped as-is. A rule-based post-processor
(\cref{fig:pipeline}) applies, in order: (1) the external Base Vegetation
Layer (BVL) mask, which limits the map to vegetated cropland; (2) a
smooth-and-reclassify step; (3) an interannual-consistency fix that
compares each pixel across years and overrides anomalous years for
stable, mostly permanent, crops; (4) a low-confidence ``undecided'' rule
applied below a probability threshold; (5) removal of the grass/fodder
class for the final product; and (6) a minimum-mapping-unit sieve that
drops objects below 0.25\,ha. The chain works, but it is brittle. Its rules are
deliberately conservative, the documentation notes that implausible
sequences still remain in the final product, and rules cannot raise
recall for rare classes or separate spectrally similar classes that
history would distinguish.
\vspace{-6pt}
\subsection{Two Sources of Free Signal}
\vspace{-3pt}
The chain exposes the two signals we bring into the model.
\vspace{-8pt}
\paragraph{Signal 1: the product's own interannual history.}
After eight cycles (2017--2024) every pixel carries two aligned yearly
sequences (one slot per production year): the past predicted class codes
and their associated confidences, where each confidence is the
probability of the predicted (winning) class, \ie the product's CTYCL
layer. We pad the sequence with no-data values out to 2029, so the same
architecture serves future years and only needs retraining on newer
interannual and BVL layers. We do not feed the model raw predictions.
The history is sampled from the interannual-consistency-fixed product
(the output of stage~3 above, keeping grass/fodder), so the model reads
the legacy rule system's own corrected history as a feature and can then
learn to go further. A pixel that was olive for seven straight years is
very different from one alternating wheat and barley, yet the baseline
never sees this.
\vspace{-8pt}
\paragraph{Signal 2: the external Base Vegetation Layer (BVL).}
The BVL is a single cropland and vegetation mask shared across all HRL
layers and produced within the consortium. It is a valid, authoritative
input that we keep using. Because the BVL is applied when the historical
product is generated, the history we sample already contains BVL-masked
observations: a pixel can appear as masked in some past years. This
raises a representation question. When a sample's own year or its history contains masked observations, how should the model read them? We study this, rather than discarding or replacing the mask.

Both signals are free, in that they need no new labels. The rest of the
paper brings them into the model: the history as an input modality, and
the masked observations as a consistently represented signal.

\section{Method}
\label{sec:method}

\begin{figure}[t]
  \centering
  \includegraphics[width=0.88\linewidth]{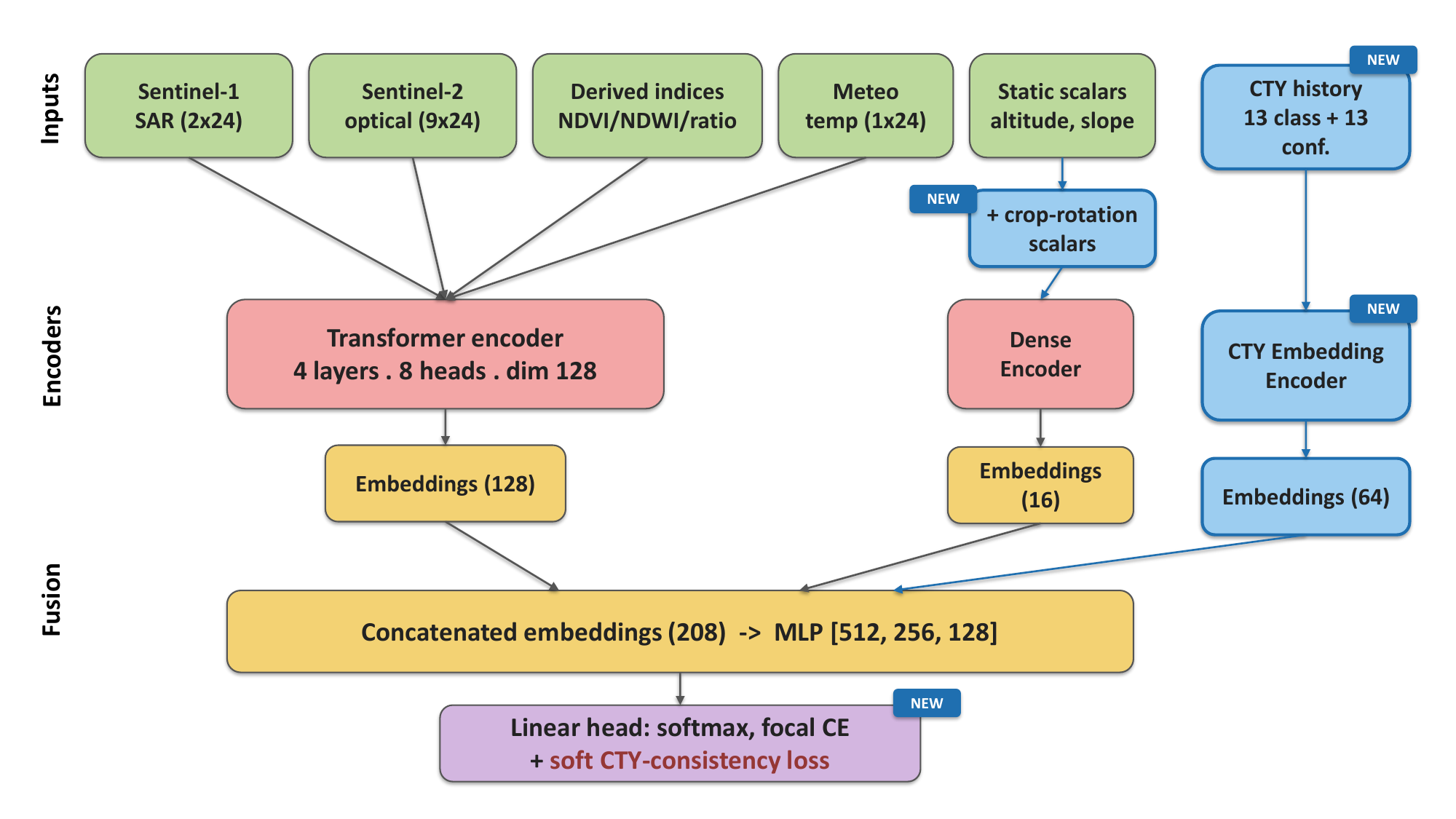}
  \caption{The multi-source architecture. The green, red, yellow, and
  purple boxes are the baseline operational model. The blue branch is our
  addition: two aligned 13-year sequences (the product's own past class
predictions and their confidences) are
  encoded by the CTY Embedding Encoder and concatenated before the MLP
  head. Crop-rotation scalars enter the dense encoder, and a soft
  consistency loss is attached to the head.}
  \label{fig:arch}
  \vspace{-3pt}
\end{figure}

\subsection{End-to-End Training instead of a Frozen Foundation Model}
\label{sec:method-rationale}
A natural question is why we train a compact, end-to-end Transformer
rather than fine-tune a pre-trained geospatial foundation model. The
answer is the data regime. Foundation models help most when labels are
scarce or when transferring to a new task; a recent deployment study on
{WorldCereal} data reaches a similar conclusion \cite{butsko2025worldcereal}.
Our setting is the opposite: our dataset has about 5.4M labelled pixels,
with good if uneven coverage across
countries and years (\cref{fig:availability}). With this many labels a
2.6M-parameter supervised model already reaches strong accuracy, while a
much larger pre-trained backbone risks overfitting in our pixel-level
setting, as the 12M-parameter variant in \cref{sec:exp-strategy}
confirms. End-to-end training also lets us learn the new history and BVL
branches jointly with the optical backbone, so the model can weigh
phenology against its own past predictions; a frozen backbone would make
that harder. We therefore keep the operational backbone and add the new
signals as trainable branches.
\vspace{-10pt}
\subsection{Baseline and Fusion}
The baseline encodes the 24-step optical, SAR, index, and meteo streams
with a 4-layer, 8-head Transformer (model dimension 128), and the static
scalars with a dense encoder. We leave this backbone untouched and add
history as a parallel encoder whose output is concatenated before the
head (\cref{fig:arch}). The design is modular on purpose: the history
branch can be added to or removed from a deployed model without
retraining the optical encoder.

\subsection{Three Ways to Encode Prediction History}
\label{sec:method-strategies}
Let a pixel's history be $\{(c_t, p_t)\}_{t=1}^{T}$ with $T{=}13$ slots
(2017--2029, future slots zero-padded), where $c_t$ is a class code and
$p_t\in[0,1]$ its winning-class confidence: two aligned sequences, $26$
values per pixel. We compare three encoders.

\noindent\textbf{(A) Scalar statistics.} The 26 raw values are fed as
continuous scalars to the dense encoder. This is simple, but it treats
nominal class codes as ordinal numbers, as if class~5 were between
classes 4 and 6, which has no meaning.

\noindent\textbf{(B) Raw time series.} The $(c_t,p_t)$ pairs are treated
as a 2-channel, 13-step series and passed to the optical Transformer.
Attention can mix history with phenology, but it still works on raw
integer codes.

\noindent\textbf{(C) CTY Embedding Encoder.} Each code is mapped
to a learnable vector, an entity embedding \cite{guo2016entity}, and
scaled by its confidence,
\begin{equation}
  \mathbf{e}_t = \mathrm{Embed}(c_t)\cdot p_t,
  \label{eq:scale}
\end{equation}
so a reliable past prediction contributes strongly and an uncertain one
contributes little. A sinusoidally initialised, learnable year-position
embedding is added, because the order of years carries meaning (a
wheat-barley sequence differs from barley-wheat). A padding-aware
Transformer attends over the valid years, followed by masked mean
pooling and a projection. The embedding table reserves index~0 as a zero
padding vector for missing or future years and, importantly for
\cref{sec:exp-bvl}, index~1 as a learnable token for externally masked
(BVL) areas. This gives a three-way distinction between no-data, masked,
and real crop. Pixels with no valid history at all (for example 2017
samples) receive a learnable ``no-info'' token, so the absence of context is represented explicitly.

\subsection{Crop-Rotation Scalars}
\label{sec:method-rotation}
We also compute two cheap scalars from the same history and route them
through the dense encoder. \emph{Crop stability} is the fraction of valid
past years that match the most recent crop, and \emph{years since change}
is the length of the current same-crop run, normalised. They give the
model a smooth, explicit summary of rotational persistence that
complements the learned embedding.

\subsection{Representing Masked Observations}
\label{sec:method-bvl}
Because the BVL mask is applied to the product, masked observations
appear both in a sample's own year and in its history, and we study how
to represent them. For the \emph{target} year, a masked observation can
be left with its crop label, given a dedicated ``masked'' class, or
folded into the existing no-crop class. For the \emph{history}, the
masked years can be left unchanged or remapped so the vector reflects
that the pixel was masked. In the history branch, the learnable BVL token
(index~1) gives past masked observations a representation that is
distinct from missing data. None of this alters or replaces the BVL; it
only decides how the model reads a signal it is already given.

\subsection{Soft Self-Supervised Consistency Loss}
\label{sec:method-loss}
To encourage the model to use a strong historical signal without leaning
on it too hard, we add a soft auxiliary term. A small linear head maps
the history embedding to class logits, with temperature scaling, and is
trained to agree with the target at a low weight ($\lambda{=}0.1$). This
rewards consistency where history is clearly informative but is too weak
to override the main focal cross-entropy. Confidence scaling, masking of the current and future years, and the weak consistency loss together act as a soft, regularizing prior.
\vspace{-8pt}
\subsection{Training}
All runs use AdamW \cite{loshchilov2019adamw}, base learning rate $10^{-3}$, focal cross-entropy,
plateau learning-rate scheduling, and early stopping on validation
macro-F1. We also ablate linear warmup and gradient clipping (norm 1.0),
which stabilise the freshly initialised embedding table early in
training. The full production model has about 2.6M parameters.
Class imbalance is handled throughout by balanced batch sampling, drawing
each batch to equalise class frequencies by target class; loss weights
are uniform unless stated otherwise.

\section{Data and Protocol}
\label{sec:data}
\begin{figure}[t]
  \centering
  \includegraphics[width=\linewidth]{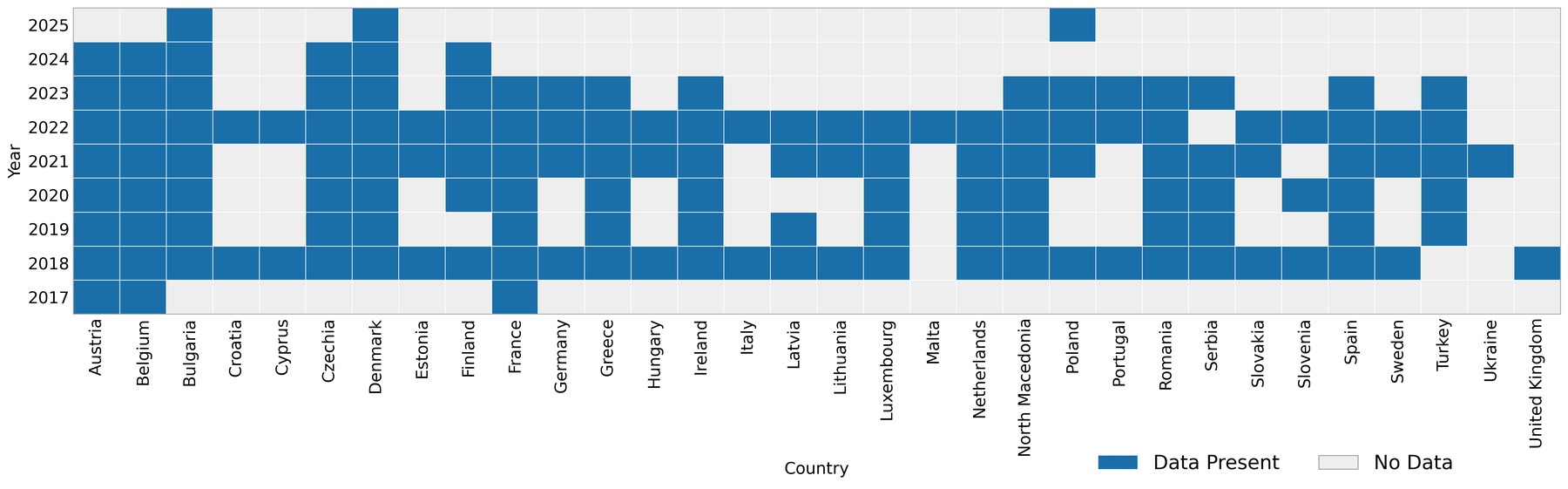}
  \caption{Reference-data availability across countries (rows) and years (columns). Coverage is uneven across space and time. However, the history input itself is dense as it comes from the prior years' model predictions, is provided up to year~$T{-}1$, while the target year~$T$ and future years are masked.}
  \label{fig:availability}
  \vspace{-12pt}
\end{figure}

\vspace{-3pt}
\paragraph{Reference data.}
Training uses a WorldCereal-harmonised \cite{vantricht2023worldcereal}
pan-European corpus of labelled pixels spanning 2018--2025 and 19 crop
classes in the product's hierarchical legend. The reference data is
pooled from national and regional sources, mostly farmer declarations
from parcel registration systems (LPIS/GSAA) plus LUCAS survey points,
and its spatio-temporal coverage is uneven: some countries have most
years, others only one or two (\cref{fig:availability}). Early
experiments use 76 such sources; later experiments add more to reach 85.
In total the corpus holds about 5.4M labelled pixels for the 85-source
set (4.8M for 76 sources). Adding newer sources made the task harder, since it brings more geographic and label diversity, and it lowers scores for older configurations. A model that matches or beats earlier numbers on the larger set is therefore genuinely better, a point we return to in
\cref{sec:exp-prod}.

\vspace{-3pt}
\paragraph{Splits.}
We split the corpus into roughly 72\% train, 13\% validation, and 15\%
test (the production run uses 3.87M train, 0.68M validation, and 0.80M
test pixels). The split is \emph{not} a random per-pixel partition: it is
stratified by crop class and grouped by field, so all pixels from one
parcel stay in the same split. This avoids the optimistic bias a random
pixel split would introduce, where neighbouring pixels of the same field
leak between train and test.

\paragraph{Inputs.}
Per pixel inputs over 24 ten-day (dekadal) windows for Sentinel-2 (9 bands), Sentinel-1 VH/VV, three derived indices (NDVI, NDWI, SAR ratio), AgERA5 temperature, along with terrain scalars, and the 13 historical CTY class
codes with their 13 CTYCL confidences (26 values). All time-domain
streams (optical, SAR, indices, meteorology) are composited to this
common 24-step, 10-day grid, so they share the same temporal sampling. A
$(0,0)$ slot is no-data; a masked observation uses the dedicated BVL
encoding.

\paragraph{Imbalance.}
The class distribution is strongly long-tailed. On the matched test
split, Grass/fodder has 319{,}359 samples and Rice only 605, a ratio of
about 528 to 1. During training this is handled by balanced batch
sampling; a further attempt to rebalance through inverse-frequency loss
weighting, applied on top of that sampling, backfires
(\cref{sec:exp-weights}).

\paragraph{Evaluation.}
We report overall accuracy, macro-F1 (the unweighted mean of per-class
F1, where F1 is the harmonic mean of precision and recall), and
weighted-F1 on held-out test splits. One caveat on comparability matters throughout. Because the
treatment of masked pixels changes the size of the no-crop class, the
test-set composition differs across configuration families; folding BVL
into no-crop, for instance, greatly enlarges that easy class. We
therefore draw all controlled encoding and optimisation comparisons
(\cref{sec:exp-strategy,sec:exp-weights}) from a single fixed split of
720{,}083 samples (the 76-source set), and we flag composition explicitly
when discussing the production evolution (\cref{sec:exp-prod}).

\paragraph{Crop-only macro score.}
To compare models whose treatment of masked pixels changes the no-crop
class, in particular the BVL variants of \cref{sec:exp-bvl}, we also
report a crop-only score: the unweighted mean of precision, recall, and
F1 over the 18 crop classes, excluding the composition-sensitive no-crop
class and any auxiliary BVL class. Because it averages per-class rates
rather than samples, this score does not depend on how non-crop pixels
are counted, and it answers the question that matters operationally:
does a design help the classes that users care about? Reporting
precision and recall separately, rather than F1 alone, shows whether a
model errs by over- or under-predicting a class.

\section{Results}
\label{sec:exp}

\subsection{Which History Encoding Wins?}
\label{sec:exp-strategy}

\Cref{tab:strategy} compares the three encoders of
\cref{sec:method-strategies} on the fixed 720{,}083-sample split. The
scalar (A) and raw time-series (B) encodings add essentially nothing over
the no-history baseline, which is expected: raw integer crop codes carry
no order for the model to use. The embedding encoder (C) gives a clear
and consistent gain and is the only single change that does so. Adding
the raw series on top (B+C) contributes a small extra signal but does not
beat C alone, and naively scaling capacity to 12M parameters hurts, a
sign that the pixel-level setting overfits easily.

\begin{table}[t]
  \caption{History-encoding comparison and training-optimisation ablation
  on the fixed test split (720{,}083 samples). Rows~1--8 compare history
  encodings; rows~9--12 (below the rule) ablate training choices on
  Strategy~C. $\Delta$ is the change in validation macro-F1 (percentage
  points) relative to the no-history baseline (row~2). All other settings
  are identical.}
  \label{tab:strategy}
  \centering
  \small
  \setlength{\tabcolsep}{4pt}
  \resizebox{\columnwidth}{!}{%
  \begin{tabular}{@{}llccccc@{}}
    \toprule
    \# & Configuration & Encoder & Params & Val F1 & Test macro-F1 & $\Delta$ \\
    \midrule
    1 & 2024 production (15 ts) & n/a & 2.55M & 0.7746 & 0.77 & $-0.30$ \\
    2 & No-history baseline & n/a & 2.55M & 0.7776 & 0.78 & n/a \\
    3 & (A) scalar stats & Dense & 2.57M & 0.7773 & 0.78 & $-0.03$ \\
    4 & (B) raw time series & Transf. & 2.78M & 0.7773 & 0.78 & $-0.03$ \\
    5 & (A+C) & Dense+Embed & 2.59M & 0.7810 & 0.78 & $+0.34$ \\
    6 & (A+C) big & Dense+Embed & 11.98M & 0.7705 & 0.77 & $-0.71$ \\
    7 & (B+C) & Transf.+Embed & 2.80M & 0.7861 & 0.78 & $+0.85$ \\
    8 & \textbf{(C) embedding} & \textbf{Embed} & 2.57M & 0.7908 & \textbf{0.79} & $+1.32$ \\
    \midrule
    9 & \textbf{(C) + warmup/clip} & \textbf{Embed} & 2.57M & \textbf{0.7916} & \textbf{0.79} & $\mathbf{+1.40}$ \\
    10 & (C) + warmup/clip + BVL-token & Embed & 2.57M & 0.7815 & 0.78 & $+0.39$ \\
    11 & (C) + inverse-freq.\ loss weights & Embed & 2.57M & 0.6439 & 0.64 & $-13.37$ \\
    12 & (C) + loss weights + BVL-token & Embed & 2.57M & 0.6306 & 0.63 & $-14.70$ \\
    \bottomrule
  \end{tabular}}
\end{table}

\subsection{Where History Helps?}
\label{sec:exp-perclass}

We break the gain down by class, comparing the baseline
with the Strategy-C model (no extra training tweaks) on the same split. The pattern is the one
multi-year reasoning predicts. The largest gains are on perennial and
tree crops with stable histories: olives +4.6, fruits +3.7, nuts +3.2,
and dry pulses +2.9 points. Abundant, spectrally distinctive classes such
as maize and sugar beet are already near their ceiling and barely move. A
field that was olive for seven straight years is almost certainly olive
again, and the embedding learns to use that. The one notable regression
is the catch-all no-crop class at field boundaries, where the history is
genuinely ambiguous.

\subsection{Training Optimisations}
\label{sec:exp-weights}

On top of Strategy~C we ablate the training choices (\cref{tab:strategy}).
Warmup with gradient clipping gives a small, consistent gain and more
stable convergence. Adding the separate BVL token inside the history
branch slightly hurts on this split, because most pixels masked in one
year are also non-crop in neighbouring years, so the token is partly
redundant with terrain features.

The clear negative result concerns class rebalancing. Every run already addresses the long tail
through \emph{balanced batch sampling}: batches are drawn to equalise
class frequencies by target class, so rare crops are well represented,
and this is our default. Adding inverse-frequency class weighting in the
focal loss \emph{on top of} this sampling rebalances the same imbalance a
second time: a rare class is then both oversampled in the batch and
up-weighted in the loss. With raw weights proportional to $N/n_c$ (a ratio
of about 528 to 1 between Rice and Grass/fodder), this double compensation
drives the model to over-predict rare classes, and macro-F1 collapses
from 0.79 to 0.64 (a 14.7-point drop in combination with the BVL token);
Grass/fodder recall falls from 92.6\% to 40.8\%.  
The production model uses uniform loss weights, and
where loss-level weighting is wanted a gentler square-root
inverse-frequency scheme avoids the collapse, as does effective-number
weighting \cite{cui2019classbalanced}.

\vspace{-8pt}
\subsection{Representing Masked Observations}
\label{sec:exp-bvl}

We now ask how the choice of representation for masked observations
affects the crop classes. We always keep the BVL mask; the question is
only how the model reads the masked observations it already sees. The
aggregate macro-F1 is misleading here, because folding masked pixels into
no-crop inflates that class (its F1 jumps to 0.92), so the headline
number rises for reasons unrelated to crop discrimination. We therefore
use the crop-only score of \cref{sec:data}, which excludes the no-crop
and any auxiliary masked class.

\begin{figure}[t]
  \centering
  \includegraphics[width=0.6\linewidth]{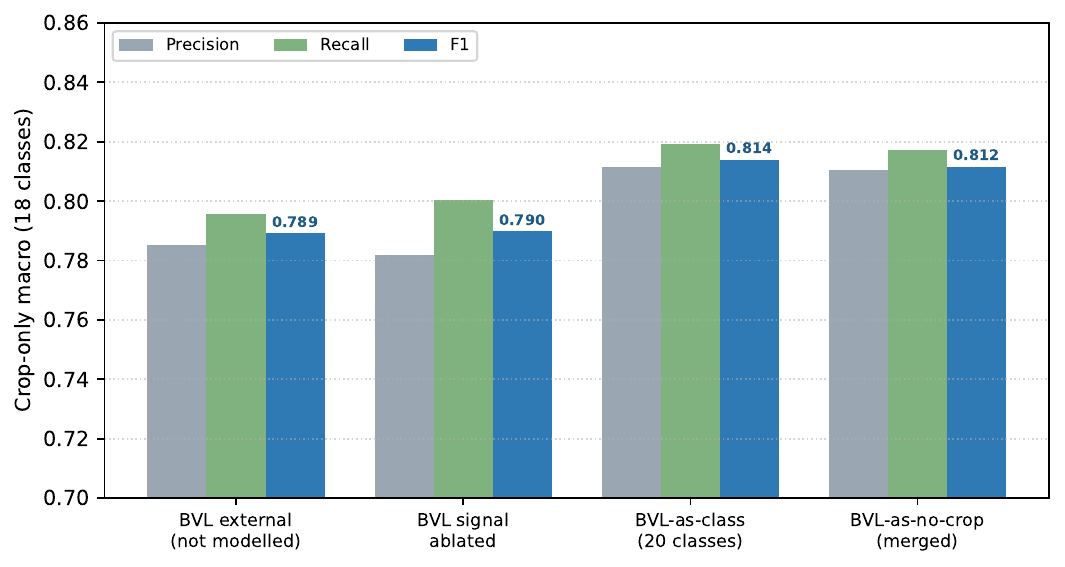}
  \caption{Crop-only macro precision, recall, and F1 across BVL
  representations. A consistent representation of masked observations
  (right two groups) 
  improves the crop classes over giving them no special
  representation 
  (left), independent of the inflated no-crop class.}
  \label{fig:bvl}
  \vspace{-12pt}
\end{figure}

\Cref{tab:bvl} and \cref{fig:bvl} agree. Giving masked observations no
special representation leaves crop-only F1 at 0.789, about the no-history
level. Representing them consistently, with a dedicated class (0.814) or
by folding them into no-crop (0.812) and remapping the masked years in
the history, raises crop-only F1 by about 2.5 points on the crop classes
themselves. The two target representations are indistinguishable on the
crops, so the choice can be made on operational grounds: folding into
no-crop keeps the 19-class schema, while a dedicated class yields an
explicit masked output. An ablation that leaves the masked years
unrepresented in the history drops crop-only F1 back to 0.790, which
shows the gain comes from representing the signal, not from the
relabelling alone.

\begin{table}[t]
  \centering
  \begin{minipage}[t]{0.55\linewidth}
    \centering
    \caption{BVL treatment compared on crop-only metrics 
    . ``Hist.'' marks
    whether the masked years are also remapped in the history; the BVL mask
    is kept in every row.}
    \label{tab:bvl}
    \vspace{2pt}
    \resizebox{\linewidth}{!}{%
    \begin{tabular}{@{}lccccc@{}}
      \toprule
      Representation of masked obs. & Hist. & Crop P & Crop R & \textbf{Crop F1} & no-crop F1 \\
      \midrule
      None (handled only by mask) & no & 0.785 & 0.796 & 0.789 & 0.55 \\
      Own-year masked dropped & no & 0.782 & 0.800 & 0.790 & 0.56 \\
      Dedicated masked class & yes & 0.812 & 0.819 & \textbf{0.814} & 0.46$^\dagger$ \\
      Folded into no-crop & yes & 0.810 & 0.817 & \textbf{0.812} & 0.92$^\dagger$ \\
      \bottomrule
    \end{tabular}}\\[2pt]
    {\scriptsize $^\dagger$no-crop F1 not comparable across rows.}
  \end{minipage}\hfill
  \begin{minipage}[t]{0.43\linewidth}
    \centering
    \caption{Crop-only macro metrics (18 classes). History improves F1 and corrects the baseline's recall
    skew (P minus R).}
    \label{tab:croponly}
    \vspace{2pt}
    \resizebox{\linewidth}{!}{%
    \begin{tabular}{@{}lcccc@{}}
      \toprule
      Model & Crop P & Crop R & \textbf{Crop F1} & P$-$R \\
      \midrule
      Baseline (no history) & 0.769 & 0.810 & 0.787 & $-0.041$ \\
      \, + History (Strategy C) & 0.806 & 0.803 & 0.803 & $+0.003$ \\
      \, + Crop-rotation + BVL-class & 0.814 & 0.823 & \textbf{0.817} & $-0.009$ \\
      \, + Consistency loss (production) & 0.810 & 0.817 & 0.812 & $-0.007$ \\
      \bottomrule
    \end{tabular}}
  \end{minipage}
\end{table}

\subsection{Precision and Recall Balance}
\label{sec:exp-pr}

Macro-F1 alone hides how a model errs, so we look at precision and recall
along the model progression (\cref{tab:croponly}). The no-history
baseline is clearly recall-skewed (P 0.769 against R 0.810, a gap of
$-0.041$): it over-predicts crops, trading precision for coverage. Adding
history raises crop-only F1 by 1.6 points and, just as usefully,
rebalances the errors (P 0.806, R 0.803). The historical signal lets the
model commit to a class with justified confidence instead of hedging.
Crop-rotation scalars and the masked-observation representation keep this balance while
lifting crop-only F1 to about 0.81.

\begin{figure}[t]
  \centering
  \includegraphics[width=\linewidth]{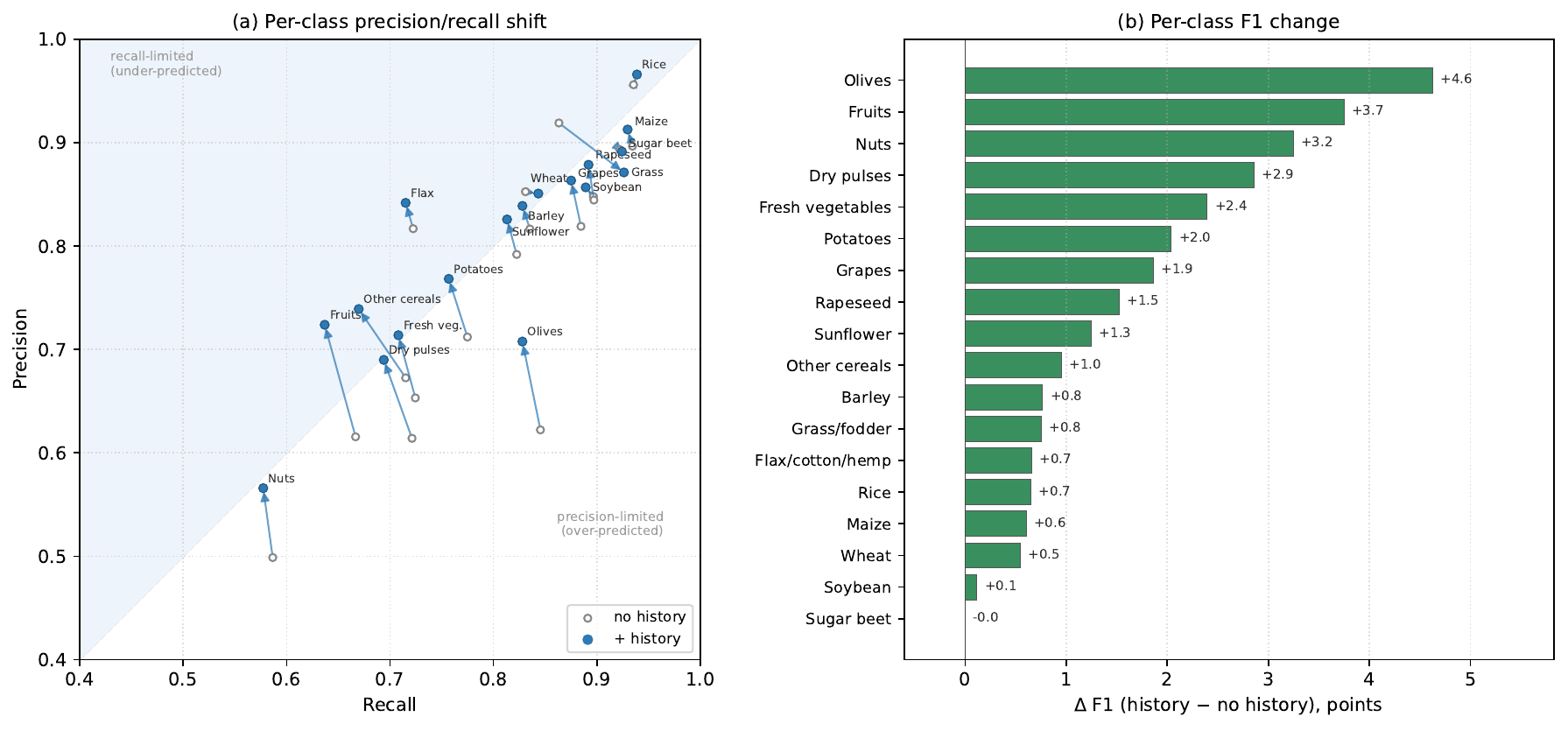}
  \caption{Effect of adding prediction history, per crop class.
  \textbf{(a)} Precision/recall shift from the no-history model (open
  markers) to the history model (filled); arrows show the movement. Most
  classes move toward the precision-equals-recall diagonal while also
  moving outward to higher values; 
  the baseline's
  over-prediction (points below the diagonal) is corrected.
  \textbf{(b)} Per-class change in F1 (history vs no-history, in points),
  sorted. Every crop class improves or holds, with the largest gains on
  perennial and tree crops (olives $+4.6$, fruits $+3.7$, nuts $+3.2$).}
  \label{fig:prshift}
  \vspace{-10pt}
\end{figure}

The question is whether history improves or worsens this balance
relative to last year's history-free model, so we trace the per-class
shift directly (\cref{fig:prshift}). Each class moves from its no-history
position (open marker) to its history position (filled), along an arrow.
The dominant motion is toward the precision-equals-recall diagonal: 13 of
the 18 crop classes become better balanced, and almost all move up and to
the right, so both metrics improve. The baseline's systematic
over-prediction, the cluster below the diagonal, is pulled back without
losing recall. The remaining structure is interpretable. Grass/fodder
stays precision-limited, as a catch-all that absorbs ambiguous pixels,
while nuts and dry pulses stay recall-limited, as rare classes the model
is cautious about. The high-support staples (wheat, maize, rice,
rapeseed) sit balanced near the top right. This points to specific
remedies, such as gentle oversampling for the recall-limited tail, far
more precisely than a single number, and it shows the gain is a real
improvement in calibration rather than a precision-for-recall trade.

\subsection{Production Model: Rotation, Consistency, and Scaling}
\label{sec:exp-prod}

The production candidate combines a deeper Strategy~C encoder (embedding
dimension 32, 2 layers, 64-d output) with the crop-rotation scalars
(\cref{sec:method-rotation}), the soft consistency loss
(\cref{sec:method-loss}), and the masked-observation representation of
\cref{sec:exp-bvl}. It reaches 0.86 accuracy, 0.82 aggregate macro-F1, and,
the metric we trust for cross-design comparison, 0.81 crop-only macro-F1
at about 2.6M parameters, with a well-balanced precision and recall
profile (\cref{tab:croponly}). It holds or improves crop-class F1 even on the harder 85-source data that lowers older configurations, the
robustness point from \cref{sec:data}. Per class, perennials again
benefit most, with olives and fruits above 0.85 and 0.73 F1.
The confusion structure is well behaved: residual errors are concentrated among agronomically similar classes (cereals; row crops) rather than spread by the historical signal.


\section{Discussion}
\label{sec:discussion}
\vspace{-10pt}
\Cref{sec:background} showed that the product already reasons across
years, but only through rolling-window diversity layers and a rule-based
interannual fix applied outside a single-year model. That post-processor
is conservative by design and cannot reason about gradual transitions.
Feeding the interannual-fixed history into the model lets it learn
class-specific persistence, gradual change, and confidence-weighted
evidence directly. Because the input is the rule-corrected product, the
model in effect distils the legacy rules and then improves on them, and
unlike the rules it can raise recall.

\begin{figure}[t]
  \centering
  \includegraphics[width=0.98\linewidth]{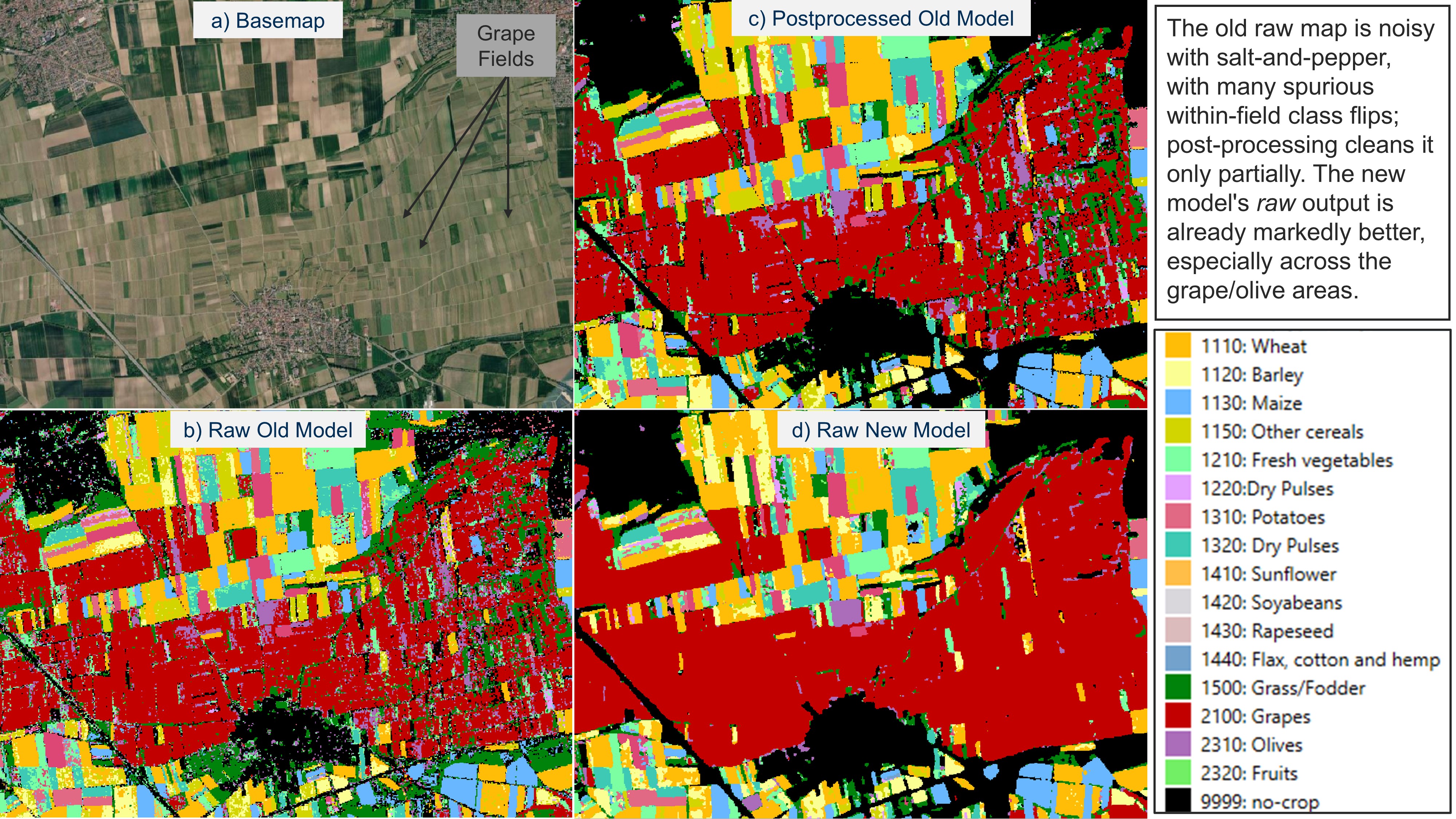}
  \caption{Inference tile sample from 2024 production model (no interannual layer) vs. from the proposed model: fewer spurious year-to-year flips and smoother field interiors. 
  }
  \label{fig:tiles}
\end{figure}

Beyond the metrics, history-aware prediction gives visibly cleaner maps.
\Cref{fig:tiles} compares inference tiles from the 2024 production model,
which has no interannual layer, with the new model. The new maps show
fewer spurious year-to-year flips on perennial parcels and smoother field
interiors, doing in one learned model what previously needed a separate
rule-based stage.

The BVL remains the authoritative mask, used as before; our only question
is how the model should read the masked observations that already appear
in its own history. As \cref{sec:exp-bvl} shows, a consistent
representation improves the crop classes (crop-only F1 from 0.79 to 0.81).
A temporally consistent model also needs fewer interannual overrides and
less sieving downstream, complementing the existing chain rather than
competing with it.

History is itself model output, so errors could in principle reinforce
themselves, and a model could learn a persistence shortcut that repeats the previous label. Several factors limit both: the current and future years are masked, so the current-year label is never an input; confidence scaling down-weights uncertain past predictions and the weak consistency loss keeps history a soft prior; and gains also appear on rotational, non-static crops, not only stable perennials. We see no runaway bias, though quantifying error accumulation over many operational cycles remains important future work. Our internal test metrics complement, rather than replace, the service's official independent validation \cite{hrlpum2025}, and tile- and parcel-level validation across held-out regions is ongoing.

\vspace{-12pt}
\section{Conclusion}
\label{sec:conclusion}
We show that a recurring crop-mapping system can be improved by feeding it signals it already produces: its interannual-fixed prediction history, encoded as confidence-scaled, time-ordered tokens, and a consistent representation of the externally provided vegetation mask. On a composition-robust crop-only score, history raises crop-only F1 by 1.6 pp, with the gains on perennial and tree crops, and rebalances a recall-skewed error profile. Representing the masked observations consistently adds about 2.5 points, and the production model reaches 0.86 accuracy and 0.81 crop-only macro-F1. More broadly, the same approach should fit any recurring Earth-observation or geospatial foundation model that emits annual or seasonal class maps, or integrates external ancillary layers in post-processing, since each production cycle builds up a prediction history that can be fed back in as a free modality or auxiliary source.

\vspace{-8pt}
\section*{Acknowledgements}
This work was carried out within the Copernicus Land Monitoring Service (CLMS)
High Resolution Layer Vegetated Land Cover Characteristics (HRL VLCC2) project,
funded by the European Environment Agency (EEA). 
The authors thank the EEA and the HRL VLCC2 consortium partners
for supporting this research.

\bibliographystyle{splncs04}
\bibliography{main}

@String(CVPR  = {IEEE Conf. Comput. Vis. Pattern Recog.})

@String(ICCV  = {Int. Conf. Comput. Vis.})

@String(NeurIPS = {Adv. Neural Inform. Process. Syst.})

@String(ICML  = {Int. Conf. Mach. Learn.})

@String(ICLR  = {Int. Conf. Learn. Represent.})

@String(CVPR  = {CVPR})

@String(ICCV  = {ICCV})

@String(NeurIPS = {NeurIPS})

@String(ICML  = {ICML})

@String(ICLR  = {ICLR})

@inproceedings{vaswani2017attention,
  author    = {Ashish Vaswani and Noam Shazeer and Niki Parmar and Jakob Uszkoreit and Llion Jones and Aidan N. Gomez and Lukasz Kaiser and Illia Polosukhin},
  title     = {Attention Is All You Need},
  booktitle = NeurIPS,
  year      = {2017}
}

@inproceedings{zerveas2021transformer,
  author    = {George Zerveas and Srideepika Jayaraman and Dhaval Patel and Anuradha Bhamidipaty and Carsten Eickhoff},
  title     = {A Transformer-based Framework for Multivariate Time Series Representation Learning},
  booktitle = {Proc. 27th ACM SIGKDD Conf. Knowledge Discovery and Data Mining (KDD)},
  year      = {2021},
  doi       = {10.1145/3447548.3467401}
}

@inproceedings{garnot2021utae,
  author    = {Vivien Sainte Fare Garnot and Loic Landrieu},
  title     = {Panoptic Segmentation of Satellite Image Time Series with Convolutional Temporal Attention Networks},
  booktitle = ICCV,
  year      = {2021}
}

@article{quinton2021crop,
  author  = {Felix Quinton and Loic Landrieu},
  title   = {Crop Rotation Modeling for Deep Learning-Based Parcel Classification from Satellite Time Series},
  journal = {Remote Sensing},
  volume  = {13},
  number  = {22},
  pages   = {4599},
  year    = {2021},
  doi     = {10.3390/rs13224599}
}

@inproceedings{tseng2024presto,
  author    = {Gabriel Tseng and Ruben Cartuyvels and Ivan Zvonkov and Mirali Purohit and David Rolnick and Hannah Kerner},
  title     = {Lightweight, Pre-trained Transformers for Remote Sensing Timeseries},
  booktitle = {NeurIPS Workshop on Tackling Climate Change with Machine Learning},
  year      = {2023}
}

@article{vantricht2023worldcereal,
  author  = {Kristof Van Tricht and Jeroen Degerickx and Sven Gilliams and Daniele Zanaga and Marjorie Battude and others},
  title   = {{WorldCereal}: a dynamic open-source system for global-scale, seasonal, and reproducible crop and irrigation mapping},
  journal = {Earth System Science Data},
  volume  = {15},
  pages   = {5491--5515},
  year    = {2023},
  doi     = {10.5194/essd-15-5491-2023}
}

@article{dandrimont2021parcel,
  author  = {Rapha\"el d'Andrimont and Astrid Verhegghen and Michele Meroni and Guido Lemoine and Pieter Defourny and Marijn van der Velde},
  title   = {From parcel to continental scale -- A first European crop type map based on {Sentinel-1} and {LUCAS} {Copernicus} in-situ observations},
  journal = {Remote Sensing of Environment},
  volume  = {266},
  pages   = {112416},
  year    = {2021},
  doi     = {10.1016/j.rse.2021.112416}
}

@inproceedings{cui2019classbalanced,
  author    = {Yin Cui and Menglin Jia and Tsung-Yi Lin and Yang Song and Serge Belongie},
  title     = {Class-Balanced Loss Based on Effective Number of Samples},
  booktitle = CVPR,
  year      = {2019}
}

@inproceedings{lin2017focal,
  author    = {Tsung-Yi Lin and Priya Goyal and Ross Girshick and Kaiming He and Piotr Doll\'ar},
  title     = {Focal Loss for Dense Object Detection},
  booktitle = ICCV,
  year      = {2017}
}

@article{russwurm2020selfattention,
  author  = {Marc Ru{\ss}wurm and Marco K\"orner},
  title   = {Self-attention for raw optical satellite time series classification},
  journal = {ISPRS Journal of Photogrammetry and Remote Sensing},
  volume  = {169},
  pages   = {421--435},
  year    = {2020},
  doi     = {10.1016/j.isprsjprs.2020.06.006}
}

@article{schneider2023eurocrops,
  author  = {Maja Schneider and Tobias Schelte and Felix Schmitz and Marco K\"orner},
  title   = {{EuroCrops}: A Pan-European Dataset for Time Series Crop Type Classification},
  journal = {Scientific Data},
  volume  = {10},
  pages   = {612},
  year    = {2023},
  doi     = {10.1038/s41597-023-02517-0}
}

@inproceedings{cong2022satmae,
  author    = {Yezhen Cong and Samar Khanna and Chenlin Meng and Patrick Liu and Erik Rozi and Yutong He and Marshall Burke and David B. Lobell and Stefano Ermon},
  title     = {{SatMAE}: Pre-training Transformers for Temporal and Multi-Spectral Satellite Imagery},
  booktitle = NeurIPS,
  year      = {2022}
}

@inproceedings{jakubik2023prithvi,
  author    = {Johannes Jakubik and Sujit Roy and C. E. Phillips and Paolo Fraccaro and Denys Godwin and others},
  title     = {Foundation Models for Generalist Geospatial Artificial Intelligence},
  booktitle = {arXiv preprint arXiv:2310.18660},
  year      = {2023}
}

@article{wang2023domain,
  author  = {Y. Wang and L. Feng and Z. Zhang and F. Tian},
  title   = {An Unsupervised Domain Adaptation Deep Learning Method for Spatial and Temporal Transferable Crop Type Mapping Using {Sentinel-2} Imagery},
  journal = {ISPRS J. Photogramm. Remote Sens.},
  volume  = {199},
  pages   = {102--117},
  year    = {2023}
}

@misc{hrlatbd2025,
  author       = {{Copernicus Land Monitoring Service}},
  title        = {{HRL VLCC} Algorithm Theoretical Basis Document (ATBD), Version 2.2},
  howpublished = {Document Code D1.2, CLMS},
  year         = {2025}
}

@misc{hrlpum2025,
  author       = {{Copernicus Land Monitoring Service}},
  title        = {{HRL} Croplands Product User Manual, Version 2.3 (CTY)},
  howpublished = {Document Code D1.12, CLMS},
  year         = {2025}
}

@article{pelletier2019tempcnn,
  author  = {Charlotte Pelletier and Geoffrey I. Webb and Fran\c{c}ois Petitjean},
  title   = {Temporal Convolutional Neural Network for the Classification of Satellite Image Time Series},
  journal = {Remote Sensing},
  volume  = {11},
  number  = {5},
  pages   = {523},
  year    = {2019},
  doi     = {10.3390/rs11050523}
}

@inproceedings{garnot2020ltae,
  author    = {Vivien Sainte Fare Garnot and Loic Landrieu},
  title     = {Lightweight Temporal Self-attention for Classifying Satellite Image Time Series},
  booktitle = {Advanced Analytics and Learning on Temporal Data (AALTD), ECML PKDD Workshop},
  year      = {2020}
}

@article{yuan2021sitsbert,
  author  = {Yuan Yuan and Lei Lin},
  title   = {Self-Supervised Pretraining of Transformers for Satellite Image Time Series Classification},
  journal = {IEEE J. Sel. Top. Appl. Earth Obs. Remote Sens.},
  volume  = {14},
  pages   = {474--487},
  year    = {2021},
  doi     = {10.1109/JSTARS.2020.3036602}
}

@inproceedings{russwurm2020breizhcrops,
  author    = {Marc Ru{\ss}wurm and Charlotte Pelletier and Maximilian Zollner and S\'ebastien Lef\`evre and Marco K\"orner},
  title     = {{BreizhCrops}: A Time Series Dataset for Crop Type Mapping},
  booktitle = {ISPRS Congress},
  year      = {2020}
}

@inproceedings{hinton2015distilling,
  author    = {Geoffrey Hinton and Oriol Vinyals and Jeff Dean},
  title     = {Distilling the Knowledge in a Neural Network},
  booktitle = {NeurIPS Deep Learning Workshop},
  year      = {2015}
}

@article{guo2016entity,
  author  = {Cheng Guo and Felix Berkhahn},
  title   = {Entity Embeddings of Categorical Variables},
  journal = {arXiv preprint arXiv:1604.06737},
  year    = {2016}
}

@inproceedings{bailly2018rotation,
  author    = {S\'ebastien Bailly and Sylvain Giordano and Loic Landrieu and Nesrine Chehata},
  title     = {Crop-Rotation Structured Classification using Multi-Source {Sentinel} Images and {LPIS} for Crop Type Mapping},
  booktitle = {IEEE Int. Geosci. Remote Sens. Symp. (IGARSS)},
  year      = {2018},
  doi       = {10.1109/IGARSS.2018.8518427}
}

@article{giordano2020rotation,
  author  = {Sylvain Giordano and S\'ebastien Bailly and Loic Landrieu and Nesrine Chehata},
  title   = {Improved Crop Classification with Rotation Knowledge using {Sentinel-1} and {-2} Time Series},
  journal = {Photogrammetric Engineering \& Remote Sensing},
  volume  = {86},
  number  = {7},
  pages   = {431--441},
  year    = {2020},
  doi     = {10.14358/PERS.86.7.431}
}

@article{abercrombie2016hmm,
  author  = {Stephen P. Abercrombie and Mark A. Friedl},
  title   = {Improving the Consistency of Multitemporal Land Cover Maps Using a Hidden {Markov} Model},
  journal = {IEEE Trans. Geosci. Remote Sens.},
  volume  = {54},
  number  = {2},
  pages   = {703--713},
  year    = {2016},
  doi     = {10.1109/TGRS.2015.2463689}
}

@inproceedings{astruc2024anysat,
  author    = {Guillaume Astruc and Nicolas Gonthier and Cl\'ement Mallet and Loic Landrieu},
  title     = {{AnySat}: One Earth Observation Model for Many Resolutions, Scales, and Modalities},
  booktitle = {IEEE/CVF Conf. Comput. Vis. Pattern Recog. (CVPR)},
  year      = {2025}
}

@inproceedings{butsko2025worldcereal,
  author    = {Christina Butsko and Kristof Van Tricht and Gabriel Tseng and Giorgia Milli and David Rolnick and Ruben Cartuyvels and Inbal Becker-Reshef and Zoltan Szantoi and Hannah Kerner},
  title     = {Deploying Geospatial Foundation Models in the Real World: Lessons from {WorldCereal}},
  booktitle = {Proc. TerraBytes {ICML} Workshop: Towards Global Datasets and Models for Earth Observation},
  series    = {Proc. Mach. Learn. Res. (PMLR)},
  volume    = {292},
  pages     = {13--31},
  year      = {2025}
}

@article{brown2022dynamicworld,
  author  = {Christopher F. Brown and Steven P. Brumby and Brookie Guzder-Williams and Tanya Birch and Samantha Brooks Hyde and Joseph Mazzariello and Wanda Czerwinski and Valerie J. Pasquarella and Robert Haertel and Simon Ilyushchenko and others},
  title   = {Dynamic World, Near Real-Time Global 10\,m Land Use Land Cover Mapping},
  journal = {Scientific Data},
  volume  = {9},
  pages   = {251},
  year    = {2022},
  doi     = {10.1038/s41597-022-01307-4}
}

@article{danish2025terrafm,
  author  = {Muhammad Sohail Danish and Muhammad Akhtar Munir and Syed Roshaan Ali Shah and Muhammad Haris Khan and Rao Muhammad Anwer and Jorma Laaksonen and Fahad Shahbaz Khan and Salman Khan},
  title   = {{TerraFM}: A Scalable Foundation Model for Unified Multisensor Earth Observation},
  journal = {arXiv preprint arXiv:2506.06281},
  year    = {2025}
}

@article{alishah2024contextual,
  author    = {Syed Roshaan Ali Shah and Obaid-Ur- Rehman and Yasir Shabbir and Rana AhmadFaraz Ishaq},
  title     = {Contextual Band Addition and Multi-look Inferencing to Improve Semantic Segmentation Model Performance on Satellite Images},
  journal   = {Journal of Spatial Science},
  volume    = {69},
  number    = {3},
  pages     = {849--872},
  year      = {2024},
  publisher = {Taylor \& Francis},
  doi       = {10.1080/14498596.2024.2305124}
}

@inproceedings{loshchilov2019adamw,
  author    = {Ilya Loshchilov and Frank Hutter},
  title     = {Decoupled Weight Decay Regularization},
  booktitle = {Int. Conf. Learn. Represent. (ICLR)},
  year      = {2019}
}

\end{document}